\documentclass[screen,nonacm,acmtog]{acmart}

\usepackage{amsmath}
\usepackage{soul}
\usepackage{bm}
\usepackage{color,balance,microtype,booktabs}
\usepackage[fleqn,tbtags]{mathtools}
\usepackage{hyperref}
\usepackage[capitalise]{cleveref} 
\usepackage[detect-weight]{siunitx}
\usepackage{pifont,xspace}
\usepackage{subfigure}
\usepackage{comment}
\makeatletter
\@namedef{ver@everyshi.sty}{}
\makeatother

\AtBeginDocument{%
  }

\newcommand{\cmark}{\ding{51}}%
\newcommand{\xmark}{\ding{55}}%

\newcommand{\name}{\textsc{LPA3D}\xspace}

\newcommand{\coordshortname}{LPA}
\newcommand{\coordfullname}{local-pose-alignment}
\newcommand{\coordanchor}{cross-scene-invariant anchors}

\begin{document}

\title{\name: 3D Room-Level Scene Generation from In-the-Wild Images}

\author{Ming-Jia Yang}
\authornote{Work done during internship at Microsoft.}
\email{yangmingjia@buaa.edu.cn}
\affiliation{%
    \institution{Beihang University}
    \city{Beijing}
    \country{P.~R.~China}
}
\author{Yu-Xiao Guo}
\email{yuxiao.guo@outlook.com}
\affiliation{%
    \institution{Microsoft Research Asia}
    \city{Beijing}
    \country{P.~R.~China}
}
\author{Yang Liu}
\email{yangliu@microsoft.com}
\affiliation{%
    \institution{Microsoft Research Asia}
    \city{Beijing}
    \country{P.~R.~China}
}
\author{Bin Zhou}
\email{zhoubin@buaa.edu.cn}
\affiliation{%
    \institution{Beihang University}
    \city{Beijing}
    \country{P.~R.~China}
}
\author{Xin Tong}
\email{xtong.gfx@gmail.com}
\affiliation{%
    \institution{Microsoft Research Asia}
    \city{Beijing}
    \country{P.~R.~China}
}

\authorsaddresses{}

\renewcommand{\shortauthors}{Yang, et al.}

\begin{abstract}
    Generating realistic, room-level indoor scenes with semantically plausible and detailed appearances from in-the-wild images is crucial for various applications in VR, AR, and robotics. The success of NeRF-based generative methods indicates a promising direction to address this challenge. However, unlike their success at the object level, existing scene-level generative methods require additional information, such as multiple views, depth images, or semantic guidance, rather than relying solely on RGB images. This is because NeRF-based methods necessitate prior knowledge of camera poses, which is challenging to approximate for indoor scenes due to the complexity of defining alignment and the difficulty of globally estimating poses from a single image, given the unseen parts behind the camera.
    To address this challenge, we redefine global poses within the framework of Local-Pose-Alignment (LPA) -- an anchor-based multi-local-coordinate system that uses a selected number of anchors as the roots of these coordinates. Building on this foundation, we introduce LPA-GAN, a novel NeRF-based generative approach that incorporates specific modifications to estimate the priors of camera poses under LPA. It also co-optimizes the pose predictor and scene generation processes. Our ablation study and comparisons with straightforward extensions of NeRF-based object generative methods demonstrate the effectiveness of our approach. Furthermore, visual comparisons with other techniques reveal that our method achieves superior view-to-view consistency and semantic normality.
\end{abstract}

\keywords{3D-aware generation; indoor scene; in-the-wild images; pose estimation}

\maketitle

\section{Introduction}

\begin{figure*}[h!t]
    \centering
    \includegraphics[width=\textwidth]{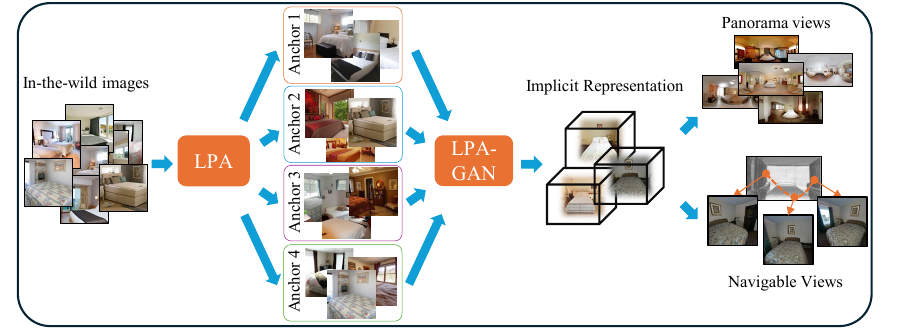}
    \caption{The workflow of \name involves a 3D room-level indoor scene generative model trained on millions of in-the-wild images with unknown global poses. This training is facilitated by local-pose-alignment, a novel pose representation that categorizes in-the-wild images into anchor-based local pose coordinate systems. Subsequently, LPA-GAN, a generative model based on neural radiance fields (NeRFs) and adapted to LPA, enables the generation of room-level scenes. It supports 360-degree panoramas and free navigation, leading to a wide range of applications.}
    \label{fig:lpa3d_workflow}
\end{figure*}

Creating realistic, room-level indoor scenes that are semantically coherent, offer a 360-degree view, and feature detailed appearance is important for applications in virtual reality (VR), augmented reality (AR), and robotics. A direct learning-based method involves collecting large-scale, real, or highly realistic 3D room-level indoor scenes to train 3D generation models. However, collecting such scenes is labor-intensive and unlikely to scale. Existing datasets, such as ScanNet~\cite{dai2017scannet}, Matterport3D~\cite{chang2017matterport3d}, 3D-Front~\cite{fu20213d}, and Structured3D~\cite{zheng2020structured3d}, contain only thousands or tens of thousands of rooms, so insufficiently capture the diversity and richness of indoor scenes. On the other hand, it is convenient to collect images, and there are numerous in-the-wild images of indoor scenes available. For example, LSUN~\cite{yu2015lsun} has millions of indoor scene images. However, these images are all single-view, with unknown camera poses, making it very challenging to train a room-level 3D indoor scene generative model using them.

A promising alternative involves neural radiance fields and generative adversarial models, which have proven effective for generating 3D objects from image collections without needing known camera poses. This methodology relies on two essential elements: defining a global coordinate system and aligning training data within it, and either presupposing or estimating camera pose distributions.
The first element leverages geometric and appearance similarities to reduce spatial variance and ambiguity, yielding realistic outcomes. The second ensures unbiased sampling, avoiding systematic errors. Extending the success of NeRF-based GANs from objects to indoor scenes is challenging. Defining a global coordinate system and aligning training data is complex due to room structures, object categories, and spatial relationships. Improper alignments increase spatial ambiguity, leading to artifacts. Acquiring prior knowledge of camera pose distribution is impractical. Camera distribution in-the-wild is unbalanced and irregular due to human activity and spatial accessibility. Typical assumptions, like uniform or Gaussian priors, are ineffective. Estimating camera poses for all images is infeasible. Precision and generality issues, along with unseen parts of the scene, complicate camera pose estimation in a global coordinate system.

To address these challenges, we draw inspiration from the human ability to rapidly approximate global camera pose from a given image. Humans identify multiple anchors within different local regions of a room, which may be distinctive objects or structural elements, and estimate the relative poses between the camera and a specific anchor. They are also aware of the strong spatial relationships between these anchors, possessing a strong prior understanding of the layout of objects and the structure of rooms designed to facilitate one or more human activities. For instance, when viewing an image of a bedroom, a person could quickly notice the position and orientation of the bed and estimate the camera pose relative to it, then, by leveraging the prior knowledge of the bed's typical placement, deduce the camera's approximate global pose.
Following this philosophy, if we can identify unique anchors with spatial relationships that remain constant across different scenes, and if we can define the local camera pose of a given image relative to its corresponding anchor, then we could approximate a global view of this image. Consequently, we could estimate prior knowledge of the global camera if the poses of all images are properly approximated.

We introduce \name, a framework designed to generate room-level 3D indoor scenes from in-the-wild images, as shown in \cref{fig:lpa3d_workflow}, which does not require camera information in the training data. To address the challenge of global pose prediction and alignment for a given image, we propose a novel multi-local-coordinate system, \emph{\coordfullname}, based on \emph{\coordanchor}. We outline the principles for selecting appropriate anchors and suggest using `core object'-based corners for them. Moreover, we introduce LPA-GAN, a NeRF-based GAN methodology for room-level 3D scene synthesis. For integration with LPA, LPA-GAN employs a boundary controller for anchor identification in generated scenes, coupled with a view-by-view matching camera sampler to emulate the global camera distribution. Most importantly, LPA-GAN implements a co-optimization strategy to concurrently train a local camera pose predictor and a scene generation model.

We have conducted an ablation study to verify the effectiveness of LPA for the generative model and the co-optimization scheme, compared to approaches with a naive extension. We also compare our work to existing methods, showing our approach surpasses them in terms of inter-view consistency and semantic reasonableness.

Our contributions are as follows:
\begin{itemize}
    \item Reformulation of the alignment and estimation of global camera poses for 3D indoor scene generation from a collection of in-the-wild images into a new definition: an anchor-based multi-local-coordinate system named LPA. This novel approach allows NeRF-based GAN methods to transition seamlessly from object-level to scene-level generation.
    \item A novel NeRF-based GAN approach that addresses both indoor scene generation and camera pose distribution reconstruction through iterative optimization, adapted to the proposed LPA. Experiments demonstrate that our generative model design surpasses naive extensions and achieves superior inter-view consistency and semantic normality
    \item To the best of our knowledge, the first method to accomplish 3D room-level scene generation trained on in-the-wild images, utilizing purely 2D RGB supervision, without the need for known camera poses.
\end{itemize}

\section{Related Work}

\subsection{Indoor scene layout synthesis}
Intrinsic relationships between objects within indoor scenes have been explored for decades. Early research utilized manually designed priors~\cite{chang2015text} or interior design principles~\cite{merrell2011interactive} for object arrangement, often with minimal or no template samples, lacking universality. With the advent of annotated 3D datasets, learning-based methods have shown their generality and quality. Diverse representations of indoor scenes include graphs~\cite{wang2019planit}, trees~\cite{li2019grains}, matrices~\cite{zhang2020deep, yang2021scene}, and images~\cite{wang2018deep,ritchie2019fast}. Advanced architectures have been developed, ranging from CNN-based to transformer-based methods~\cite{wang2021sceneformer, paschalidou2021atiss, para2023cofs}. Notably, LEGO-Net~\cite{wei2023lego} and DiffuScene~\cite{tang2024diffuscene} drew inspiration from diffusion models, approaching indoor scene synthesis as an object-set denoising process. However, these methods heavily rely on 3D datasets, posing acquisition and scalability challenges. Recent efforts aimed to reduce this dependency by using 2D-only guidance to train a 3D layout synthesizer, utilizing resources like depth-semantic images~\cite{yang2021indoor} or known-pose semantic images~\cite{nie2023learning}. Another approach involved procedural layout generation with priors from pretrained large language models~\cite{fu2024anyhome, feng2024layoutgpt, aguina2024open}. These methods share a common limitation: the output of simple 3D semantic bounding boxes, followed by object retrieval from databases, leading to a lack of realistic appearance and failure to support various downstream tasks. A recent study~\cite{fang2023ctrl} introduced a post-processing step using a diffusion model to generate realistic panoramic images from semantic boxes, but this technique still required paired data of 3D semantic annotations and panoramic images for training. Such data is scarce and difficult to acquire. Consequently, this research area continues to rely significantly on the availability of annotated 3D data, which is limited and unlikely to match the vast scale of images needed to replicate the success of generative vision models.

\subsection{3D scene generation from image diffusion models}
Stable Diffusion~\cite{rombach2022high} has demonstrated impressive visual quality, having been trained on millions or billions of images. Fine-tuning techniques like ControlNet~\cite{zhang2023adding} provide robust and flexible control over generation, whether prompted by text, semantic or depth images, or masked color images, heralding a new era of indoor scene generation. RGBD2~\cite{lei2023rgbd2} undertook scene completion through novel view inpainting from multiple-view images with known poses. Text2Room~\cite{hollein2023text2room} addressed indoor scene generation through incremental generation—starting with view inpainting from text and known parts, progressing through color image to 3D mesh, and culminating with mesh blending. Additional efforts have applied this concept to various 3D representations, such as NeRFs~\cite{zhang2024text2nerf} and 3D Gaussian splatting~\cite{chung2023luciddreamer}. However, these approaches, as pointed out by the RoomDreamer work~\cite{song2023roomdreamer}, suffer from significant artifacts when dealing with conflicts between view-view generation. A potential solution is to train a multi-view diffusion model to generate multiple overlapping views simultaneously, with notable works including RoomDreamer, DiffCollage~\cite{zhang2023diffcollage}, and MVDiffusion~\cite{tang2023mvdiffusion}. Unfortunately, due to the requirement of multi-view images as input, these diffusion methods cannot train with numerous images, leading to an overall quality drop. Some post-optimization methods, like 3D-SceneDreamer~\cite{zhang20243d}, ShowRoom3D~\cite{mao2023showroom3d}, and ControlRoom3D~\cite{schult24controlroom3d}, resolved fine-level visual quality with an extra image diffusion refinement step. Compared to our method, the intrinsic limitation of this line of work is that it often produces globally implausible layouts which do not match normal room layouts (see further discussion in \cref{sec:results}), despite their realistic details and appearancer.

\subsection{3D-aware generative adversarial network}
Beyond the achievements of generative adversarial networks in image generation~\cite{karras2019style, sauer2022stylegan}, researchers are continually seeking methods to create content that is 3D-aware or maintains 3D consistency, using only images for supervision. DepthGAN~\cite{shi20223d} presented a joint RGB-depth generation framework, where depth is estimated from an off-the-shelf pretrained depth prediction model, incorporating a rotation consistency constraint to facilitate 3D-aware generation. Meanwhile, BlobGAN~\cite{epstein2022blobgan} and BlobGAN3D~\cite{wang2023blobgan} employed sparse latent information to implicitly control scene consistency when the viewpoint changed. However, these methods cannot precisely guarantee 3D consistency due to the lack of an explicit 3D representation. Neural radiance field-based generative methods, such as PiGAN~\cite{chan2021pi}, EG3D~\cite{chan2022efficient}, and EpiGRAF~\cite{skorokhodov2022epigraf}, show promising results in object-level generation with the simple presupposition of prior knowledge of camera distribution. Considering the complexity of object-object relationships and the flexibility of camera poses, extending NeRF-based GAN from objects to scenes is not straightforward. This extension leads to additional requirements beyond mere RGB images lacking known poses. GSN~\cite{devries2021unconstrained} utilized multi-view RGB-D images as input, assuming a uniform distribution of poses. Subsequently, GAUDI~\cite{bautista2022gaudi} further improved the input by incorporating captured camera poses with dense RGB-D sequences for enhanced visual results. SinGRAF~\cite{son2022singraf} utilized multi-view RGB images without known camera poses but was confined to a single room. Conversely, DisCoScene~\cite{xu2023discoscene} and CC3D~\cite{bahmani2023cc3d} required a semantic layout as prior knowledge. All these studies impose strict requirements on the input images, which are difficult to satisfy for large-scale, in-the-wild images---a challenge our work seeks to address.

\subsection{Camera prior learning}
Several research initiatives have suggested self-learning camera priors via a co-optimization strategy. An initial study, MP-GAN~\cite{li2019synthesizing}, utilized a specialized camera predictor in conjunction with a generative model, iteratively refining both networks for a specific object category using silhouette images. PoF3D~\cite{shi2023learning} presented a technique for automatically optimizing the camera pose with two degrees of freedom (DoF) for in-the-wild RGB images. Subsequently, 3DGP~\cite{skorokhodov20233d} and GET3D~\cite{yu2023get3d} broadened flexibility of camera pose to six DoF. TEFF~\cite{chen2024learning} enhanced quality by implementing a matching loss between features of views processed by a pretrained vision foundation model, thus improving camera pose estimates. These studies concentrated on object-level generation, relying on the considerable variation in object projection from different perspectives. However, this concept is less relevant to indoor scenes, where the presence of multiple objects and a broader range of viewpoints within a single room call for more intricate methods, as proposed in our research.

\section{Method}

\subsection{Cross-scene-invariant anchors based \coordshortname{}}
\subsubsection{Concepts}
Instead of estimating the global pose of a given image and constructing camera distribution statistics from all images, we propose a novel solution, \coordfullname{}. We begin by introducing the reformulated problem and its core component, \coordanchor{}. Subsequently, we present a novel approach to label anchor identification for in-the-wild images. Finally, we conduct an analysis to demonstrate the robustness of our labeling method and the posterior knowledge on the imbalanced distribution of views.

\subsubsection{Problem reformulation}
Cross-scene-invariant anchor based LPA addresses the problem by assuming that if a multi-local-coordinate system is definable, and the spatial relationships between the anchors---which act as the roots of the local coordinates---remain relatively stable across various scenes, then it is possible to approximate the global pose of a single image. This approximation is achieved by determining the specific anchor with which the given image is associated and estimating its corresponding pose within the local coordinate system.

Two principles are crucial for selecting or defining anchors for LPA:
\begin{itemize}
    \item The definition must be three-dimensional and consistent across scenes.
    \item Each anchor should be distinctive and easily identifiable or computable for both real and synthetic images within a typical generative framework.
\end{itemize}
The first principle ensures that the spatial relationships between anchors remain nearly constant across scenes, which is essential for using multi-local coordinates to approximate those in the global coordinate system. The second principle emphasizes practicality, as without a simple or robust method to identify anchors, the accuracy of the estimated poses cannot be guaranteed. Anchors that meet these two criteria are referred to as \coordanchor{}.

To adapt the generative model for in-the-wild images, three steps are necessary:
\begin{enumerate}
    \item Selecting \coordanchor{}.
    \item Defining the local coordinate system for each anchor.
    \item Devising a method to identify these anchors and estimate the local pose relative to the corresponding anchor in both real images and synthesized scenes.
\end{enumerate}

Once the LPA has been defined, for any given real or synthesized image, we represent its camera pose, $\Phi$, as an 8-DoF vector $\{a, x, y, z, \alpha, \beta, \gamma, f\}$. Here, ${a}$ denotes the ID of the selected anchor, $\{x, y, z\}$ specifies the local position, and $\{\alpha, \beta, \gamma\}$ defines the orientation in the corresponding anchor's local coordinate system. Lastly, considering the various sensors used to capture images in the wild, we incorporate the field of view, denoted $f$, into the coordinate system. We assume an aspect ratio of \num{1} and that the image is square for simplicity.

\begin{figure}[t!]
    \centering
    \includegraphics[width=\linewidth]{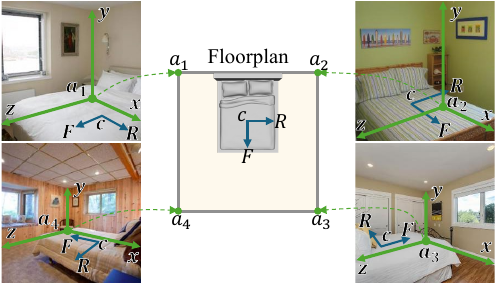}
    \caption{
        Process of anchor identification and establishing a local coordinate system based on the corners of a `core object' $c$, such as a bed in a bedroom. Given an image, the front ($F$) and right-side ($R$) directions of $c$, are determined. The corner is then identified, and its ID is queried based on predefined criteria, like $a_1$, located at the left side of the bed's head as shown in the floorplan. Finally, a local right-hand coordinate system is constructed with axes $\{\bm{x}, \bm{y}, \bm{z}\}$.
    }
    \label{FIG:AnchorDesc}
\end{figure}

While various options for selecting the \coordanchor{} exist, we have chosen the corners where walls meet floors in a room. This choice assumes that all rooms are cuboids, featuring exactly four corners where each wall meets the floor. For each local coordinate, we use a right-hand coordinate system, with the $y$-axis always on top. Additionally, we identify these anchors by their spatial relationship to the `core object' in each room type. We manually assign a `core object' for each room type: a bed in bedrooms, a sofa in living rooms, and a stove in kitchens. \cref{FIG:AnchorDesc} illustrates a typical example of anchor identification and local coordinate establishment using `core object'-based corners.

There are several advantages to defining such `core object'-based corners as coordinate anchors. Firstly, the definition of wall-floor corners is stable across scenes, and the spatial relationships between all corners are solely related to the prior boundaries of rooms, which are easy to input as a conditional signal. Secondly, the corner is conducive to inference or calculation for both real and synthetic images.

For synthetic images, since the camera pose and boundaries are known, all corners can be reverse-projected into the screen space with depth information, making it easy to determine the corresponding corner. Additionally, calculating the local camera pose relative to the corresponding anchor is straightforward, owing to the global knowledge of both the camera's and the anchor's definitions. For real images, although there is no method to determine location of a specific corner, we have found that humans are adept at this identification task by considering the relationships between the `core object' and corners, even when the corners are obscured by other furniture. Furthermore, we also note that the task is highly reliable and robust with a classifier trained even with a small subset of images (see  \cref{Sec:AnchorIdentification}). Intrinsic and extrinsic properties of real images can be inferred either from pretrained models or through a co-optimization approach with a generative model. The latter is more challenging, yet it offers a choice that we can select. We offer a comparison in the results section to illustrate which method yields the optimal outcome.

In any given image, multiple corners may be visible, or none at all. For simplicity and without loss of generality, we select the leftmost corner as the responsible anchor when multiple corners are present. If no corner is visible, we categorize it as the closest left corner outside the screen.

\begin{figure*}[!t]
    \centering
    \includegraphics[width=\textwidth]{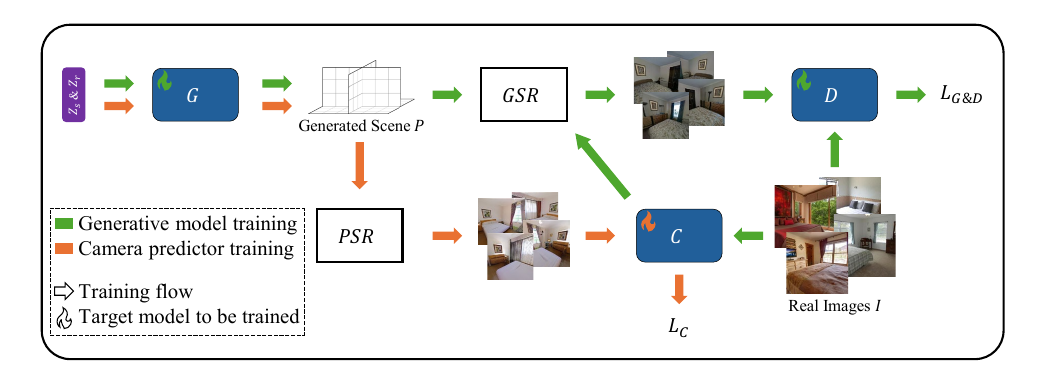}
    \caption{
        Overview of LPA-GAN. We adopt a co-optimization strategy to iteratively train the generative model and camera predictor. During the training iterations of the generative model, highlighted by green lines, only the \emph{generator} ($G$) and \emph{discriminator} ($D$) are trainable. The \emph{camera predictor} ($C$), which remains frozen during this phase, estimates camera poses from real images, providing these poses as candidates for subsequent sampling processes. The \emph{generative sampler for rendering} ($GSR$) then selects a camera pose within the specified synthetic scene to render a synthetic view. In contrast, during the training iterations of the \emph{camera predictor}, depicted by orange lines, the \emph{predictive sampler for rendering} ($PSR$) uniformly samples views within synthetic scenes generated by the now-frozen \emph{generator}. $PSR$ supplies rendered images and their corresponding camera poses as training samples for training the \emph{camera predictor}.         }
    \label{fig:Overview}
\end{figure*}

\subsubsection{Anchor identification for in-the-wild images}\label{Sec:AnchorIdentification}
The definition of `core object'-corner based \coordanchor{} allows for rapid identification of the corresponding anchor through simple multi-class classification labeling. People can make accurate decisions quickly due to the distinct orientation of the `core object' and the significantly different surrounding contents. However, annotating millions of images remains labor-intensive. We propose training a classifier with a few manually labeled images to ensure robustness and generality. First, the content is distinguishable among different anchors due to the differences in content and the `core object.' Second, instead of training a classifier from scratch, which could be sensitive to appearance changes and lead to poor generality, we utilize a pretrained segmentation backbone InternImage~\cite{wang2023internimage} to extract features. We then add six convolutional layers with global pooling to obtain the logits for prediction. This design minimizes the interference of appearance, making the predictions robust and accurate, even with a very small subset of images as training data.

In-the-wild images display significant non-uniformity across anchors, as shown later in \cref{FIG:CornerDistribution}, reflecting the highly non-uniform distribution of camera poses. This data imbalance can result in suboptimal generation quality in sparsely observed regions. To mitigate this challenge, we propose a simple yet effective strategy: equalize the sample count for each anchor to match the maximum observed number by repetitively tiling the samples. An ablation study, detailed in \cref{Sec:DataBalance}, substantiates the efficacy of this data balancing strategy.

\subsection{LPA-GAN framework}

\subsubsection{Overview}

Our method takes a latent code $z_s$, sampled from a Gaussian distribution $Z_s$, and a room size vector ${x_r, y_r, z_r} \in z_r$, sampled from a prior of room sizes $Z_r$, as inputs. We use a collection of in-the-wild images as training data, denoted $I$, where each image is captured with unknown camera poses, encompassing both intrinsic and extrinsic parameters. We assume that each image is captured independently, in the sense that there is no guarantee that there are multiple views of the same scene. For each image, the camera pose is represented by an 8-DoF vector ${a, x, y, z, \alpha, \beta, \gamma, f} \in \Phi$, where the anchor ID $a$ is predicted using the method given in \cref{Sec:AnchorIdentification}, while the remaining parameters are unknown. We aim to generate complete room-level 3D indoor scenes, represented by an implicit neural radiance field $P$. To align these synthetic scenes, we define the center of the room's floor as the origin of the radiance field coordinate system. To accommodate varying sizes of synthetic scenes and ensure consistent physical object sizes across scenes, we establish a predefined maximum room size that can be represented by the implicit radiance field $P$ and maintain it unchanged across all synthetic scenes. For different scenes, we use the room size vector $z_r$ to control their boundaries within this maximum size.

An overview of our design is given in \cref{fig:Overview}. Our framework adopts a typical neural radiance field-based generative model, which includes a generator $G$ to generate the implicit radiance field $P$ of a scene, and a discriminator $D$ to evaluate the views rendered from the radiance field $P$. We utilize the tri-plane representation, as proposed by EG3D, for the radiance field $P$. This representation uses three $256 \times 256$ planes with \num{32} channels to capture the $xyz$-dimensions in the 3D feature space, and we follow their network architecture for both $G$ and $D$. To tackle the complex distribution of camera poses, we introduce a boundary-aware generator in \cref{SubSec:BrdG}, which governs the synthesis scene boundaries and provides a valid boundary for camera sampling within the synthesized rooms. Subsequently, we introduce an LPA-based mechanism that approximates this process in real-time, comprising three components: a camera predictor $C$, a generative sampler for rendering $GSR$, and a predictive sampler for rendering $PSR$. The camera predictor $C$ estimates the 8-DoF camera pose $\Phi$ for a given image, as described in \cref{SubSec:CamPred}. The $GSR$ identifies the optimal camera pose for rendering a view by utilizing a set of candidate poses and the given radiance field $P$ of a synthetic scene, as explained in \cref{SubSec:GSR}. The $PSR$ supplies pairs of images and camera poses as training samples for the camera predictor, generated from synthetic scenes, as discussed in \cref{SubSec:PSR}. Finally, the detailed workflow to train both the generative model pipeline and the camera predictor pipeline, along with the optimization target, is explained in \cref{SubSec:Opt}.

\subsubsection{Boundary-aware generator}\label{SubSec:BrdG}
To effectively synthesize scenes with a given boundary as the condition, we propose using a room-size conditional generator. This generator assigns content based on the given room size $z_r$ as a prior. Additionally, we introduce a view-based boundary alignment loss, which encourages the alignment of the generated content with the boundary.

We adopt a method similar to SGSDI for transforming a parametric room size $z_r$ into a dense grid, which is subsequently expanded into multiple scaled feature maps via an encoder. These features are then injected into each layer of the generator at the corresponding resolutions. The primary distinction lies in our use of a 2D grid, as opposed to their 3D grid, reflecting our particular choice of representation. This method ensures thorough incorporation of boundary information during generation.

To ensure correspondence between the generated content and the specified boundary, we adhere to a straightforward yet effective principle: the fundamental structure of rooms, such as floors, walls, and ceilings, should be affixed to the boundary, whereas other objects should be contained within it. Utilizing an off-the-shelf pretrained segmentation backbone InternImage, we are able to classify and pinpoint as background the pixels $P_B$ that constitute the basic structure of a room, and designate the remaining ones, $P_F$, as foreground. Subsequently, we introduce a content-aware boundary loss, $L_B$, to impose constraints separately on the foreground and background:
\begin{align}
    L_B = \frac{1}{N} \left(\sum_{p\in{P_F}} \max(d_p - d_b, 0) + \sum_{p\in{P_B}} {\lVert {d_p - d_b} \rVert_1}\right).
\end{align}
Here, $N$ represents the total number of pixels. The predicted depth of a pixel, $d_p$, is calculated by integrating the depth with density. The desired depth, $d_b$, is determined from the camera to the boundary.

\subsubsection{Camera Predictor}\label{SubSec:CamPred}

We propose a dedicated camera predictor to estimate the camera pose $\Phi$ from any given image, which is crucial for determining the distribution of camera poses from in-the-wild images. We believe that pose information is more closely related to semantics rather than the color and texture of an RGB image. Therefore, we extract semantic content features from the multiscale output features of the encoding phase from InternImage and use a lightweight convolutional network to aggregate and process these multiscale features into the predicted camera pose $\Phi$. Specifically, we add convolution layers with a stride of 2 and a kernel size of 3 to downsample the features, aligning them with the resolutions of the coarsest-level features. We then sum these features pixel-wise, flatten the output, and use a linear layer to align the output with the camera pose $\Phi$.

\subsubsection{Generative sampler for rendering}\label{SubSec:GSR}

A straightforward approach would be to statistically analyze the distribution of camera poses across all training samples and then sample a camera pose for generated scenes based on these statistics. However, since the camera predictor is trained with synthetic scenes that evolve with the generator, this method is inefficient and unstable.

Instead, we use a view-by-view matching strategy to query an identical view of the generated scene from a camera pose estimated from a real image. This method is more flexible and robust during co-optimization, ensuring a good approximation after several epochs of training. However, there may be significant differences in the detailed layout between the generated scene and the real image. For instance, a camera pose derived from the real image could become invalid if the estimated position places the camera within an object in the generated scene, making direct migration of the camera pose infeasible.

To address this, we select multiple real images and estimate their camera poses as candidates for each view to be generated. For each candidate camera pose $\Phi_i$, we calculate its density $\rho_i$ at its global position ${x^g_i, y^g_i, z^g_i}$, derived from its local position ${x_i, y_i, z_i}$ relative to anchor $a_i$ and the absolute position ${x^a, y^a, z^a}$ of anchor $a_i$ in the radiance field $P$ of the generated scene. We then apply a softmin function to these densities $\rho_i$ to determine the sampling probability for these candidates. A candidate $\Phi_i$ is then randomly selected according to its probability. Regions occupied by furniture typically exhibit higher densities, and cameras falling into high-density areas are less likely to be sampled. This approach effectively reduces invalid poses while maintaining the stochastic nature of the camera selection process.

\subsubsection{Predictive sampler for rendering}\label{SubSec:PSR}

We utilize synthesized scenes to train the camera predictor, where camera poses are uniformly sampled within the boundary. To avoid the sampled position ${x,y,z}$ falling into objects, we employ a strategy similar to the generative sampler for the rendering module, which samples multiple candidates and selects one based on densities. For ${\alpha,\beta,\gamma}$, we uniformly sample the yaw $\alpha \in [0^{\circ}, 360^{\circ}]$, the pitch $\beta \in [-10^{\circ}, 40^{\circ}]$, and the roll $\gamma \in [-5^{\circ}, 5^{\circ}]$. Once the position and rotation of the camera pose have been selected, the anchor $a$ is determined by reversing the projection of all anchors back to the view space and selecting the anchor based on visibility. We then render the corresponding color image, pass it through the camera predictor, and supervise the predicted camera pose $\Phi'$ with the ground truth $\Phi$ using the camera prediction loss $L_C$:
\begin{align}\label{EQ:CELoss}
    L_C = \sum_{c\in\Phi, c'\in{\Phi'}} H(c,c'),
\end{align}
where $H(\cdot,\cdot)$ is the cross-entropy loss.

\subsubsection{Joint Optimization}\label{SubSec:Opt}

We introduce a joint optimization approach between the proposed generative model and the camera predictor. This approach aims to enhance scene synthesis, which in turn improves camera prediction. Better camera prediction subsequently leads to higher generative quality, resulting in a more precise approximation of the camera distribution. Thus, we iteratively optimize the generative model, including both the generator $G$ and discriminator $D$, and the camera predictor $C$ during training iterations.

For the generative model optimization pipeline, we iteratively optimize the generator $G$ and discriminator $D$. For the generator $G$, the pipeline first synthesizes a radiance field $P$ from sampled latent variables $z_s$ and room size $z_r$. It then selects the most appropriate camera pose $\Phi$ from a dozen camera poses predicted by the frozen camera predictor from randomly selected real images. With the radiance field $P$ and the selected camera pose $\Phi$, a synthesized image can be rendered and passed through the frozen discriminator, obtaining the loss of the generative model $L_G$, defined as below:
\begin{align}
    L_G = L_\mathrm{Adv}^G + \lambda_B L_B + \lambda_C L_C^D,
\end{align}
where $L_\mathrm{Adv}^G$ is the non-saturating GAN loss for the generator, $L_B$ is the boundary loss, and $L_C^D$ is the camera reconstruction loss introduced by GRAM~\cite{deng2022gram} to enhance discriminative capability by incorporating both color and camera information. Empirically, we set $\lambda_B = \lambda_C = 1$. For the discriminator, we start with a randomly selected image and processes it through the discriminator $D$, with the loss of the discriminative model $L_D$ defined by:
\begin{align}
    L_D = L_\mathrm{Adv}^D + \lambda_\mathrm{R1} L_\mathrm{R1} + \lambda_C L_C^D + \lambda_K L_K,
\end{align}
where $L_\mathrm{Adv}^D$ is the non-saturating GAN loss for the discriminator and $L_\mathrm{R1}$ is the R1 gradient penalty. We also use the camera pose reconstruction loss $L_C^D$ and a knowledge distillation loss $L_K$ from 3DGP. We set $\lambda_\mathrm{R1} = \lambda_C = \lambda_K = 1$.

For the camera predictor optimization pipeline, we randomly synthesize a radiance field $P$ using the frozen generator and select an optimal camera pose $\Phi$ from a batch of randomly generated camera poses. We then apply the camera prediction loss $L_C$ as the optimization goal.

\section{Experiments} \label{sec:results}

\subsection{Experimental setup} \label{subsec:setup}

\subsubsection{Dataset}
We selected three image categories—\emph{bedroom}, \emph{living room}, and \emph{kitchen}—from the LSUN dataset, the most extensive indoor image repository, to serve as our in-the-wild training dataset. There were approximately 3 million bedroom, 1.3 million living room, and 2.2 million kitchen images. We adhered to the common practice of filtering out low-quality images through instance selection~\cite{devries2020instance} and removing images containing watermarks and borders, reducing these numbers of training images to 1.1 million, 0.45 million, and 0.82 million respectively.

\subsubsection{Dataset for anchor identification} For each room dataset, we randomly selected 10,000 images and labeled them with one of four types of anchors to train the anchor classifier described in \cref{Sec:AnchorIdentification}. The labeling process was efficient, requiring only visual assessment, taking half a day for a person to manually label one dataset.

\subsubsection{Training setup}
Our network was implemented in PyTorch and trained using the Adam optimizer. For the GAN network, the learning rates were set to $0.0001$ for the camera predictor and $0.002$ for both the generator and the discriminator. Training was conducted on eight NVIDIA Tesla 16GB V100 GPUs, with a batch size of 128 and 100,000 iterations. Our model was trained for each room type individually, with the training time spanning one week.

\subsubsection{Network size and efficiency}
Our generator and discriminator respectively had 36.7 million and 27.9 million network parameters, comparable to  EG3D. The  camera predictor had 22.0 million trainable parameter parameters, and its pretrained backbone contained 80 million parameters. Inference time for the generator was \SI{20}{ms}, using \SI{2.2}{GB} peak GPU memory.

\begin{figure*}[!t]
    \centering
    \includegraphics[width=\textwidth]{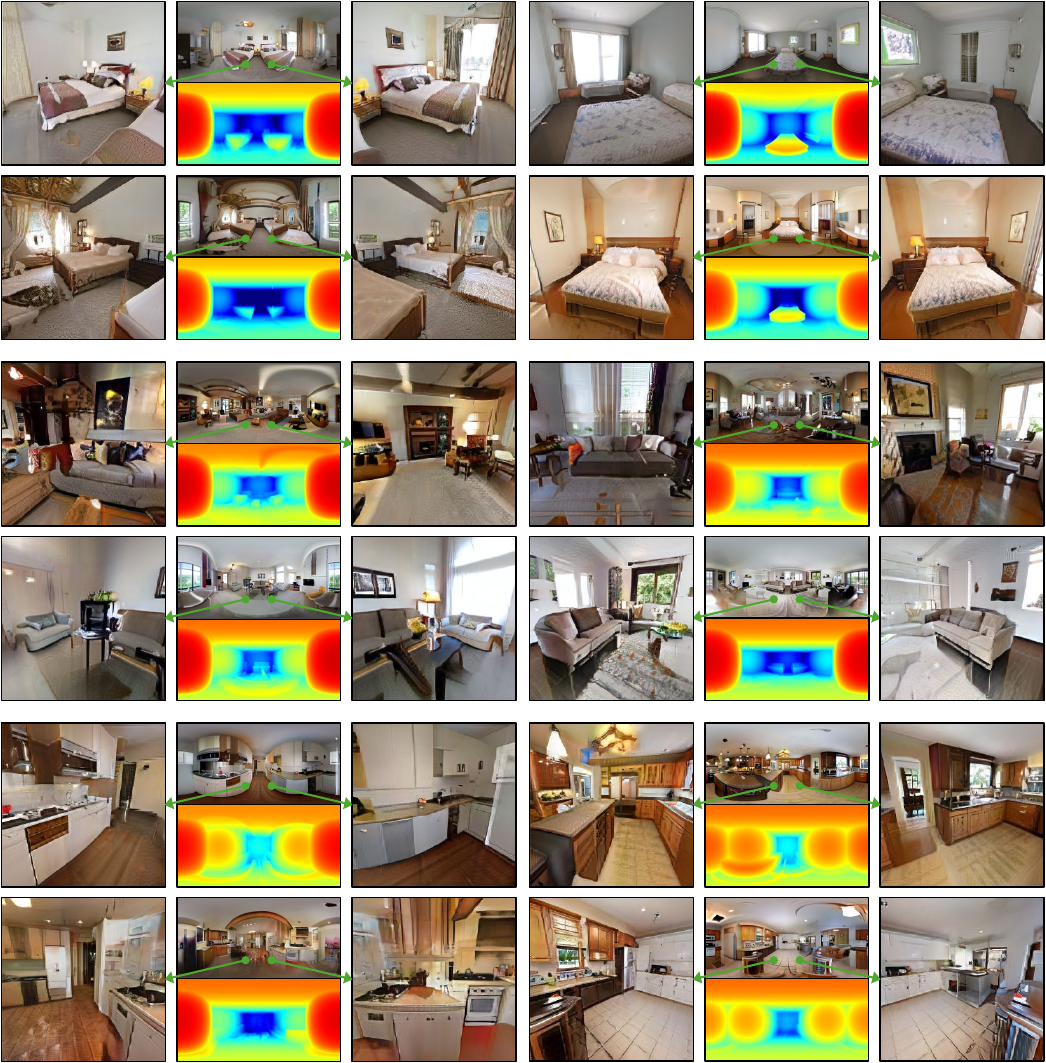}
    \caption{Various 3D scenes generated by our model. The room types, arranged from top to bottom, are bedroom, living room, and kitchen. Each scene is depicted as a panoramic view and two perspective views, including both RGB and depth images. }
    \label{FIG:Visual}
\end{figure*}

\subsubsection{Evaluation metrics}
We employed two perceptual metrics to assess generation quality. The first was the Fréchet inception distance (FID) score~\cite{Heusel_Ramsauer_Unterthiner_Nessler_Hochreiter_2017}, which measures the disparity between the distribution of images in the training dataset and the distribution of rendered images from the generated 3D scenes. For each room dataset, we generated 50,000 3D scenes and rendered each scene from four random camera positions sampled by our camera sampler.

\begin{figure*}[!t]
    \centering
    \begin{minipage}[t]{0.05\linewidth}
        \centering
        \rotatebox{90}{\ \ \ \ \ \ DepthGAN}
    \end{minipage}
    \begin{subfigure}{}
        \includegraphics[width=0.9\linewidth]{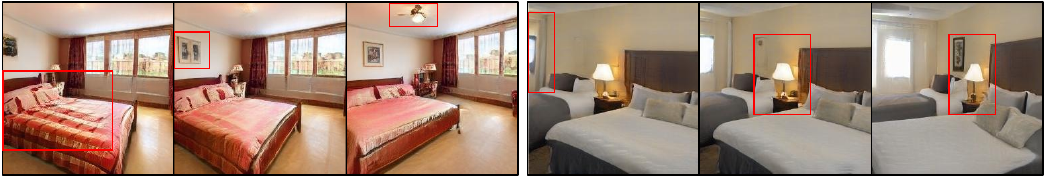}
    \end{subfigure}
    \quad
    \\
    \begin{minipage}[t]{0.05\linewidth}
        \centering
        \rotatebox{90}{\ \ \ \ BlobGAN-3D}
    \end{minipage}
    \begin{subfigure}{}
        \includegraphics[width=0.9\linewidth]{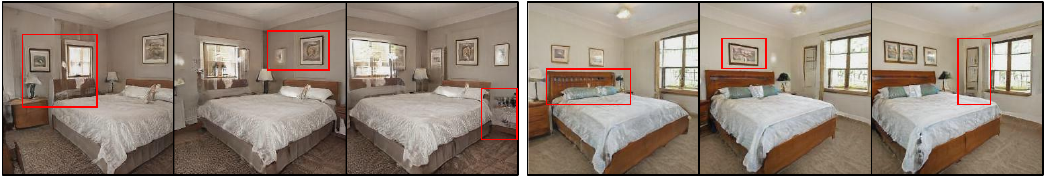}
    \end{subfigure}
    \quad
    \\
    \begin{minipage}[t]{0.05\linewidth}
        \centering
        \rotatebox{90}{\ \ \ \ \ \ \ \ \ EG3D}
    \end{minipage}
    \begin{subfigure}{}
        \includegraphics[width=0.9\linewidth]{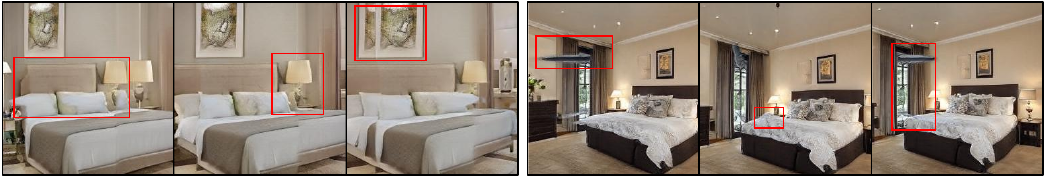}
    \end{subfigure}
    \quad
    \\
    \begin{minipage}[t]{0.05\linewidth}
        \centering
        \rotatebox{90}{\ \ \ EG3D-Camera}
    \end{minipage}
    \begin{subfigure}{}
        \includegraphics[width=0.9\linewidth]{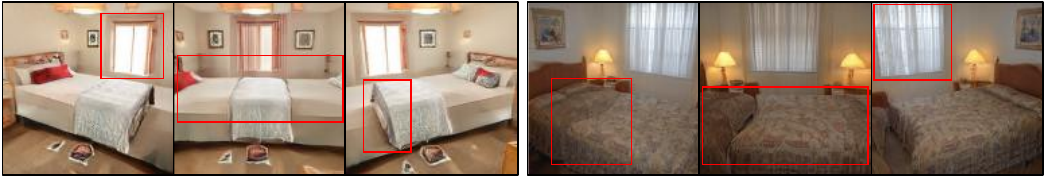}
    \end{subfigure}
    \quad
    \\
    \begin{minipage}[t]{0.05\linewidth}
        \centering
        \rotatebox{90}{\ \ \ \ \ \ Text2Room}
    \end{minipage}
    \begin{subfigure}{}
        \includegraphics[width=0.9\linewidth]{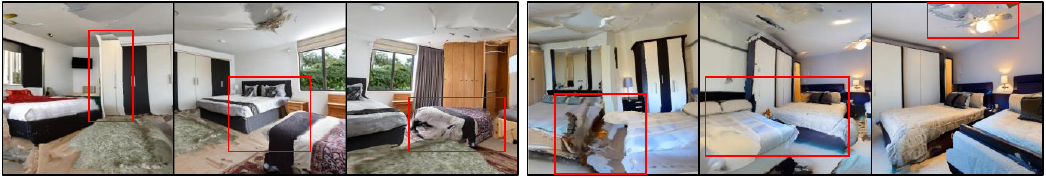}
    \end{subfigure}
    \quad
    \\
    \begin{minipage}[t]{0.05\linewidth}
        \centering
        \rotatebox{90}{\ \ \ \ \ \ \ \ \ \ \ \ Ours}
    \end{minipage}
    \begin{subfigure}{}
        \includegraphics[width=0.9\linewidth]{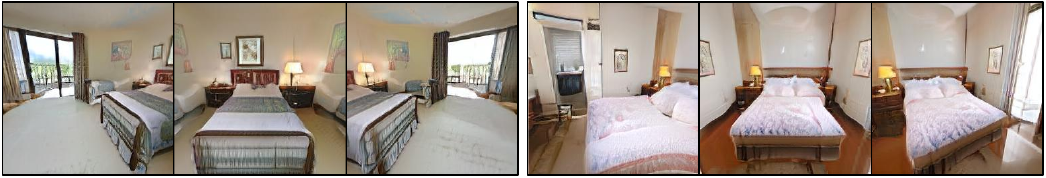}
    \end{subfigure}
    \quad
    \\
    \caption{Qualitative comparisons with existing indoor scene generation methods.
        Each row contains two scenes, each rendered from three different views. The view-inconsistent 3D contents and abnormal layout arrangement are highlighted using red boxes.
    }
    \label{FIG:Comparison}
\end{figure*}

The second metric measured whether the generated layout contradicts common room design; for instance, it is unlikely for a bedroom to have two beds opposite each other. We refer to this metric as \emph{layout abnormality}. Layout abnormality can be easily detected by viewers but it is challenging to define comprehensively and to measure robustly and automatically. For simplicity, we randomly generated 50 bedrooms and manually counted the occurrence rate of two opposed beds, and used this rate to define layout abnormality and evaluate different approaches to bedroom generation. Other types of layout abnormality can be seen in \cref{FIG:Comparison} later.

Additionally, we conducted a user study to compare the perceptual quality of scenes generated by our method to those from other methods, as described later.

\subsection{Model evaluation} \label{subsec:eval}

\subsubsection{Qualitative and quantitative evaluation}
We present several 3D scenes generated by our model in \cref{FIG:Visual}. For each scene, we display a rendered panoramic view and two rendered perspective views, as well as the depth map. Additionally, videos demonstrating navigation through these generated scenes are available in the Supplementary Material. The generated scenes are visually plausible, showing diverse room layouts and objects, such as beds, sofas, and kitchen tables.
The FID scores of our model are $12.7$ (bedroom), $18.6$ (living room), $14.0$ (kitchen).
For bedroom generation, the layout abnormality score is $5.7\%$.

\subsubsection{Comparison}
We compared our model to (i) 3D-aware generative models: DepthGAN, BlobGAN-3D, EG3D and EG3D-Camera (which used a given camera prior in the global coordinate system to generate complete 3D scenes) trained on the bedroom images of the LSUN dataset, and (ii) diffusion-based 3D generative models such as Text2Room which leveraged pretrained text-to-image diffusion models. Since the BlobGAN-3D code is not publicly available, we used the examples provided in their paper for comparison. To extend EG3D from face generation to indoor scene generation, we trained EG3D on the LSUN dataset with two different camera priors. The first camera prior was consistent with the camera prior in face generation, following the setting of EG3D re-implemented by BlobGAN-3D, generating only local regions of indoor scenes. The second camera prior was defined in the global coordinate system of indoor scenes, uniformly rotated 360 degrees around the yaw axis $\beta$, generating a complete 3D indoor scene.

\cref{FIG:Comparison} presents the 3D scenes generated by our method alongside these approaches. As indicated by red boxes, BlobGAN-3D and DepthGAN struggle to generate view-consistent 3D content, as they are essential 2D-based methods and do not address view inconsistency. Thus view-inconsistent contents are frequently observed (such as fans, lights, wall paintings, pillows and nightstands in the figure). Although EG3D is a 3D-based method, the camera in EG3D always focuses on the local area of the scene, resulting in a biased prior of camera distribution. Consequently, EG3D generates scenes with abnormal geometry. The manually provided camera prior in EG3D-Camera often fails to match the camera prior of the training data, compromising the generation quality of EG3D-Camera and leading to a high layout abnormality score in the generated scenes. Text2Room always generates view-consistent content as its 3D scene representation is a 3D mesh; however, its iterative inpainting process only ensures local layout normality, and a lack of awareness of global layout normality. As the figure shows, Text2Room's results may include unusual bed arrangements, such as  lifted beda, beda on the ceiling, and two close and oppositely-arranged beds, which are rarely seen in everyday life or in the common image dataset.

We also conducted a user study to compare scenes generated by EG3D-Camera, Text2Room, and our method. In this study, we presented 50 pairs of generated scenes to 12 participants, with each pair containing one scene generated by our method and one by another method. Participants were asked to select the scene they perceived to have a more reasonable layout.
As \cref{TAB:UserStudy} shows, the participants exhibited a clear preference for the scenes generated by our method over those produced by the other methods.

\begin{table}[t!]
    \centering
    \caption{In the user study, participants preferred scenes generated by our method over those generated by EG3D-Camera and Text2Room.}
    \begin{tabular}{lcc}
        \toprule
        \textbf{Comparator} & \textbf{ Ours preferred} & \textbf{Other preferred} \\ \midrule
        EG3D-Camera         & \textbf{78.7\%}          & 21.3\%                   \\
        Text2Room           & \textbf{97.3\%}          & 2.7 \%                   \\
        \bottomrule
    \end{tabular}
    \label{TAB:UserStudy}
\end{table}

Additionally, navigation videos within these generated scenes are available in the Supplementary Material, which distinctly highlight the deficiencies of the scenes, including view inconsistency, abnormal geometry and abnormal layout.

\subsubsection{Model scalability}
As we train our model with in-the-wild images, it is essential to explore the scalability of our model concerning the volume of training data. We assessed this scalability by training the model across various data scales: 5,000, 20,000, 100,000, and 1,000,000 bedroom images, each randomly selected from the LSUN dataset. \cref{TAB:DataScaleAblation} presents the FID scores for each corresponding model. The results clearly demonstrate a consistent enhancement in model performance as the training data scale increases, underscoring our model's potential to effectively leverage larger datasets of in-the-wild images.

\begin{table}[t!]
    \centering
    \caption{Model scalability with respect to the training data size.}
    \label{TAB:DataScaleAblation}
    \begin{tabular}{lcccc}
        \toprule
        \textbf{Training data size} & \SI{5}{k} & \SI{20}{k} & \SI{100}{k} & \SI{1}{M} \\ \midrule
        {FID}                       & $46.0$    & $21.5$     & $17.2$      & $12.7$    \\
        \bottomrule
    \end{tabular}
\end{table}

\subsubsection{Anchor identification robustness \& quality}
\begin{figure}[t!]
    \centering
    \includegraphics[width=\linewidth]{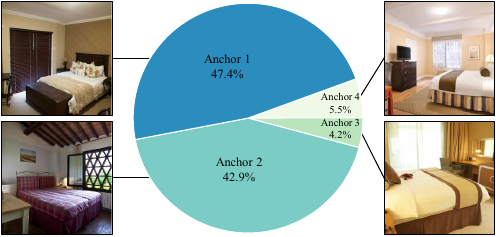}
    \caption{Occurrences of images belonging to each anchor in the LSUN-Bedroom dataset.}
    \label{FIG:CornerDistribution}
\end{figure}

We evaluated the robustness and quality of the proposed classifier, which is trained on a small subset of images and assists in automatically labeling the remaining unlabeled ones. To this end, we set up experiments to train the classifier with varying numbers of labeled images and used an additional set of 2,000 images for testing, with both training and test examples selected from the LSUN-Bedroom dataset. We measured the mean accuracy across different anchors of images, as shown in \cref{Table:AccWithImages}. The results indicate that the task of anchor identification is not overly difficult and achieves high accuracy even with a few thousand training images, considering that the total number of candidate images is 3 million.
The image counts for the identified anchor types are visualized in \cref{FIG:CornerDistribution}, revealing an uneven distribution.

\begin{table}[t!]
    \centering
    \caption{Mean accuracy of anchor identification using a varying number of labeled images as training data, across 2,000 test images from the LSUN-Bedroom dataset.}
    \label{Table:AccWithImages}
    \begin{tabular}{c|cccc}
        \toprule
        \textbf{Num. of labeled images} & 1,000 & 2,000 & 4,000 & 8,000 \\
        \midrule
        {Accuracy $\%$}                 & 82.7  & 86.9  & 90.9  & 94.0  \\
        \bottomrule
    \end{tabular}
\end{table}

\begin{figure}[t]
    \centering
    \includegraphics[width=\linewidth]{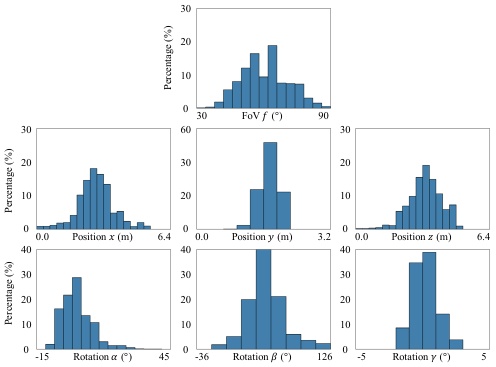}
    \caption{The distribution of camera parameters predicted by the online trained camera predictor.}
    \label{FIG:LSUNCamera}
\end{figure}

\begin{figure}[t]
    \centering
    \includegraphics[width=\linewidth]{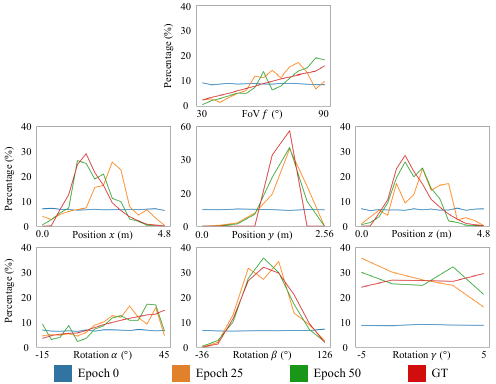}
    \caption{Camera parameters estimation on the Structured3D-Bedroom dataset. The figure illustrates the histogram distribution of estimated camera parameters of the training images and their ground-truth, as well as our intermediate prediction results at the beginning (Epoch 0), middle (Epoch 25) and the end (Epoch 50) of our training process. }
    \label{FIG:CameraShifting}
\end{figure}

\subsubsection{Camera pose estimation}
Our model is designed to progressively improve camera pose estimation during training, without requiring prior knowledge of ground-truth camera poses. In \cref{FIG:LSUNCamera}, we present histograms of the estimated camera pose information for the training images, which exhibit an uneven distribution of camera poses. Since the LSUN dataset lacks ground-truth camera poses for quantitative comparison, we conducted another experiment to evaluate the accuracy of our model. We chose bedroom images from the Structured3D dataset for our training set. Structured3D comprises approximately 4,500 bedrooms created by professional artists, including perspective and panoramic images, along with ground-truth camera parameters. We extracted 100,000 square images from these bedrooms and resized them to a resolution of $256 \times 256$. We further processed these images with our corner alignment network and transformed the ground-truth camera parameters into a local coordinate system for evaluation purposes.

We trained our model on these images without using any camera information. After training, we used the trained camera predictor to estimate the camera parameters for all the training images. \cref{FIG:CameraShifting} shows histograms of both predicted and ground-truth camera parameters, and \cref{TAB:CameraShifting} reports the mean absolute error between the predicted and ground-truth camera parameters, as well as the intermediate prediction results during training. We observed that our predictions gradually converged towards the ground-truth camera distribution as training progressed, demonstrating the effectiveness of our joint learning scheme.

\begin{table}[t]
    \centering
    \caption{Mean absolute errors of camera pose estimation on the Structured3D-Bedroom dataset, at the beginning (Epoch 0), middle (Epoch 25) and the end (Epoch 50) of our training process.}
    \label{TAB:CameraShifting}
    \begin{tabular}{lrrr}
        \toprule
        \textbf{Camera}                        & Epoch 0   & Epoch 25  & Epoch 50  \\ \midrule
        \text{Position} $x$ (m)                & $\;\,1.3$ & $\;\,0.7$ & $\;\,0.5$ \\
        \text{Position} $y$ (m)                & $\;\,1.2$ & $\;\,0.2$ & $\;\,0.2$ \\
        \text{Position} $z$ (m)                & $\;\,1.3$ & $\;\,0.6$ & $\;\,0.5$ \\
        \text{Rotation} $\alpha$ (\textdegree) & $34.2$    & $\;\,4.3$ & $\;\,3.6$ \\
        \text{Rotation} $\beta$ (\textdegree)  & $71.6$    & $15.8$    & $14.5$    \\
        \text{Rotation} $\gamma$ (\textdegree) & $\;\,9.6$ & $\;\,2.0$ & $\;\,1.6$ \\
        \text{FoV} $f$ (\textdegree)           & $17.3$    & $\;\,9.7$ & $\;\,7.5$ \\
        \bottomrule
    \end{tabular}
\end{table}

\subsection{Ablation Study}\label{Sec:Ablation}

We conducted the following ablation studies to validate our design choices.

\subsubsection{Camera Pose Prior}
In our work, we treat the camera poses of the training data as unknowns and learn them in conjunction with 3D generation. An alternative and intuitive approach is to assume a distribution of camera poses and utilize it for GAN training, as in many 3D-aware object generation works such as GSN. We incorporated this prior into our network, which we term \emph{PriorCam}: $f$ was sampled from a normal distribution (mean = $60^\circ$, standard deviation = $15^\circ$), and $x, y, z$ were uniformly sampled within the room box. The angles $ \alpha, \beta, \gamma$ were uniformly sampled from the ranges $[0^\circ, 360^\circ]$, $[-10^\circ, 40^\circ]$, and $[-5^\circ, 5^\circ]$ respectively. As reported in \cref{TAB:Ablation}, PriorCam performed poorly due to a significant mismatch between its camera pose prior and the actual camera poses.

\begin{table}[t!]
    \centering
    \setlength\tabcolsep{0pt}
    \caption{Comparisons of different training configurations on LSUN bedroom under $256\times 256$ resolution. The FID and the abnormality score are reported.
    }
    \label{TAB:Ablation}
    \begin{tabular*}{\columnwidth}{@{\extracolsep{\fill}} lrrrr}
        \toprule
        \textbf{Configuration} & \textbf{Pretrain} & \textbf{Training Data} & \textbf{FID}$\downarrow$ & \textbf{Abnormality}$\downarrow$ \\ \midrule
        PriorCam               & Use Prior         & Balanced               & $53.0$                   & $23.3\%$                         \\
        ScratchPredictor       & \xmark            & Balanced               & $17.6$                   & $17.3\%$                         \\
        OriginalData           & \cmark            & Original               & $18.0$                   & $32.7\%$                         \\
        AnchorFree             & \cmark            & Original               & $16.4$                   & $72.7\%$                         \\
        Default                & \cmark            & Balanced               & $12.7$                   & \;\,$5.7\%$                          \\
        \bottomrule
    \end{tabular*}
\end{table}

\subsubsection{Effects of Pretraining}
In our experiments, we found that using a pretrained network as the backbone to extract image features for camera pose prediction is essential for improving network efficacy. We attempted training without using pretrained weights, termed \emph{ScratchPredictor}, and observed that the training became unstable and the FID increases from $12.6$ to $17.6$.

\subsubsection{Training data balance}\label{Sec:DataBalance}

As revealed by \cref{FIG:CornerDistribution}, the number of images falling into different anchor-based local coordinate systems is unbalanced. For instance, in the LSUN-bedroom dataset, there is a strong preference for photos taken from the foot of the bed, facing forward. If this unbalanced behavior is not addressed and the network is trained directly on the data, many camera poses will be undersampled, degrading the generation quality. We trained our network without data balance, termed \emph{OriginalData}, and confirmed its inferior performance, as shown in \cref{TAB:Ablation}.

\subsubsection{Anchor Type Discrimination}
\begin{figure}[t!]
    \centering
    \includegraphics[width=\linewidth]{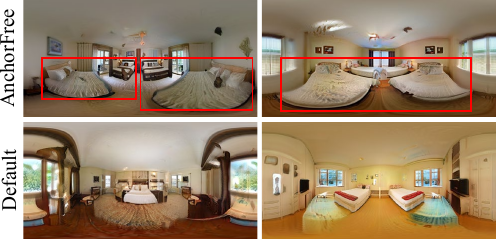}
    \caption{Qualitative comparison between the \emph{AnchorFree} and \emph{Default} configurations. Multiple beds are generated along adjacent walls, an arrangement that is abnormal for a standard bedroom layout.}
    \label{FIG:Ablation}
\end{figure}

Our camera-aware discriminator loss (\cref{EQ:CELoss}) includes a loss term $H(a^{D'},a^D)$ that penalizes incorrect predictions of anchor types. This loss is crucial for ensuring that the generated images adhere to the distribution of the training images with the same anchor type. In an ablation study in which we removed the use of anchors and this loss from our model, termed \emph{AnchorFree}, we found that the generated scenes were mostly unreasonable. This is evidenced by the highlighted regions in \cref{FIG:Ablation}, despite a marginal decrease in the FID score.

\subsection{Limitations}

\begin{figure}[t]
    \centering
    \includegraphics[width=\linewidth]{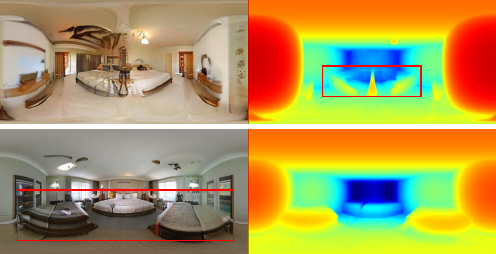}
    \caption{Two generated instances exemplifying limitations of our proposed method. In the first scene, the surface of the bed exhibits a collapse, and the geometry is inconsistent with the ground truth. In the second scene, an anomalous configuration is observed where three beds are positioned within a bedroom, deviating from conventional spatial arrangements.}
    \label{FIG:FailureCases}
\end{figure}

There are a few limitations in our work that require further study. Firstly, the current room layout presumes a box-like structure, which does not account for the varied layouts of many actual rooms, thus violating our assumption. Incorporating a diversity of room layout patterns into our anchor-based local coordinate system would help enhance our model training.
Secondly, our approach exclusively utilizes 2D RGB images for training, which may lead to discrepancies in the geometry of the generated scenes relative to the ground truth, as exemplified in the first scene of \cref{FIG:FailureCases}. Leveraging existing depth estimation techniques, which readily extract depth cues from RGB images, could provide depth priors which could enhance our scene generation model's capability to reproduce scenes with accurate geometry.
Lastly, scenes generated by our method occasionally exhibit abnormal global layouts, as observed in the second scene of \cref{FIG:FailureCases}. This limitation stems from the fact that the perspective image in the training data only captures a local area of the entire scene. Introducing panoramic views or 3D layout data could serve as global layout supervision, thereby enhancing the plausibility of the overall scene layout.

\section{Conclusions} \label{sec:conclusion}
In this work, we addressed the challenge of generating 3D indoor scenes from in-the-wild images, a task complicated by unknown camera poses. We proposed a novel anchor-based local coordinate system for aligning these images and predicting view poses through an effective joint-training strategy. This strategy progressively improves camera pose estimation within the local coordinate system and predicts NeRF-based 3D scenes via an adversarial generative network. Our approach ensures inter-view consistency and exhibits significantly lower layout abnormality than other methods. The validated scalability of our model positions it as a promising method to enhance 3D scene generation using large datasets of in-the-wild images.

\appendix
\subsection*{Declaration}
LPA3D is purely a research project. Currently, we have no plans to incorporate LPA3D into a product. We will also put Microsoft AI principles into practice when further developing the model.

\bibliographystyle{ACM-Reference-Format}

\end{document}